\documentclass[conference]{IEEEtran}
\usepackage{times}
\usepackage[numbers]{natbib}
\usepackage{multicol}
\usepackage{amsmath,amssymb}
\usepackage{booktabs}
\usepackage{graphicx}
\usepackage{xcolor}
\usepackage{tikz}
\usetikzlibrary{positioning,arrows.meta}
\usepackage{pgfplots}
\pgfplotsset{compat=1.17}
\usepackage[bookmarks=true]{hyperref}
\hypersetup{pdfauthor={Ji Pei},pdftitle={Imagination Needs the Right History: World-Action Models versus VLAs for Long-Horizon Dexterous Bimanual Manipulation},hidelinks}

\newcommand{\vla}{\textsc{Q-VLA}}
\newcommand{\wam}{\textsc{Q-WAM}}
\newcommand{\mm}{\,mm}
\newcommand{\finding}[1]{\textbf{#1}\;}

\title{Imagination Needs the Right History:\\
World-Action Models versus VLAs for\\
Long-Horizon Dexterous Bimanual Manipulation}

\author{\IEEEauthorblockN{Ji Pei$^{*}$}
\IEEEauthorblockA{Xi'an Jiaotong University, Xi'an\\ $^{*}$\texttt{ji.pei@stu.xjtu.edu.cn}}}

\begin{document}
\maketitle

\begin{abstract}
World-action models (WAMs) predict future video together with actions, on the
premise that a policy which imagines the future acts better. We test whether
this foresight survives deployment, on a nine-stage bimanual routine with two
7-DoF arms and two 20-DoF hands, learned from 2{,}932 glove-teleoperated
demonstrations. A flow-matching VLA and a video-diffusion WAM are trained on
the same data with the same action space, served through the same interface
and compared offline on held-out episodes. Deployment takes three things from
the WAM. \emph{History:} with the dense visual history it was trained on, the
WAM is 1.9\mm\ more accurate than the VLA, and a control trained without video
prediction attributes 0.8\mm\ of this to imagining the future; with the one
frame per request that a robot actually sends, the lead is gone. \emph{Time:}
scored on the actions executed after each policy's measured latency (100
vs.\ 308\,ms), the VLA is 4.0\mm\ more accurate, and the WAM stays ahead only in
keeping a resting hand still. \emph{Continuity:} both policies jump about
19\mm\ where consecutive action chunks meet, three times the natural per-frame
motion. We turn these observations into four falsifiable hypotheses and a
paired, operator-blind real-robot protocol.
\end{abstract}

\IEEEpeerreviewmaketitle

\section{Introduction}

A single image tells a robot hand where the shaker is, but not where in its
swing the shaker is. WAMs make the missing information explicit: they predict
future video together with actions, so that the physical priors of large video
generators~\cite{wan2025} shape control~\cite{du2023unipi,wu2024gr1,li2025uva,zhu2025uwm,hu2025vpp}.
Recent work scales the visual context of these models~\cite{huang2026longwam}
and compares their design choices under shared
protocols~\cite{yu2026openwam,wang2026openwam}. These studies ask how a WAM
should be built. We ask \emph{whether its foresight survives deployment on a
long-horizon dexterous task, and whether it then beats a strong VLA trained on
exactly the same data}.

Answering this requires holding everything but the policy fixed. Published
WAM--VLA comparisons rarely do: the baselines are reported numbers on shared
benchmarks, trained with their own action spaces, history lengths and
inference schedules~\cite{yu2026openwam,huang2026longwam}. On a bimanual
system with 54 actuated DoF the action parameterization alone (absolute or
relative arm targets, joint or end-effector space, the treatment of an idle
arm) changes what is learnable. We therefore build a controlled testbed: one
dataset, one 60-D action space, one serving interface and one stage scheduler,
behind which either a $\pi_{0.5}$-based VLA~\cite{pi05} (\vla) or a
Wan2.2-based WAM~\cite{wan2025,huang2026longwam} (\wam) runs. A third model,
\wam\ trained without video prediction, isolates what imagination contributes.

Offline, on 147 held-out episodes, deployment takes three things from the WAM.

\emph{History.} With the 1.6\,s of dense visual history it was trained on,
\wam\ is 1.9\mm\ more accurate than \vla\ in end-effector position, and the
model without video prediction attributes 0.8\mm\ of this lead to imagination.
A deployed robot, however, sends one new frame per request. Under this sparse
history the WAM trained on dense history loses 7.5\mm; fine-tuning on sparse
history recovers most of the loss and leaves the two policies within 0.5\mm\
of each other, except on
muddling and shaking, whose phase cannot be read from frames 0.67\,s apart.

\emph{Time.} A request takes 308\,ms for \wam\ and 100\,ms for \vla. Scored on
the actions that are executed once the reply has arrived, \vla\ is 4.0\mm\ more
accurate, while at equal delay the two stay within 0.5\mm\ of each other.
\wam's remaining advantage, a 29\% lower hand-joint error, turns out to be a
steadier resting hand: while the hand moves the lead shrinks to 9\%, and at
the measured latency it reverses.

\emph{Continuity.} Consecutive action chunks of either policy disagree by about
19\mm\ where they meet, three times the natural per-frame motion, and training
\wam\ on the actions already committed does not reduce the jump.

Imagination thus needs the right history and enough time to use it. Since
action error is not success rate, we state what these observations imply as
four hypotheses with a pre-specified real-robot protocol
(Sec.~\ref{sec:hyp}).

\textbf{Contributions.}
\begin{itemize}
\item \textbf{A controlled WAM--VLA comparison for dexterous manipulation.} Two
large pretrained policies share data, split, a 60-D action space,
normalization, serving interface and stage scheduler; a WAM trained without
video prediction isolates the effect of imagination.
\item \textbf{Deployment-aware offline evaluation.} Three measurements that
follow the deployment interface: error under the history a robot can send,
error of the actions executed after the measured end-to-end latency, and the
jump where chunks meet. A hold-still baseline separates holding from grasping.
\item \textbf{Findings and falsifiable hypotheses.} The WAM's lead depends on
dense history and partly on video prediction; latency outweighs it; what
remains is a steadier resting hand. Each finding becomes a real-robot
hypothesis with a pre-specified paired test and its controls.
\end{itemize}

\begin{figure*}[t]
\centering
\includegraphics[width=\textwidth]{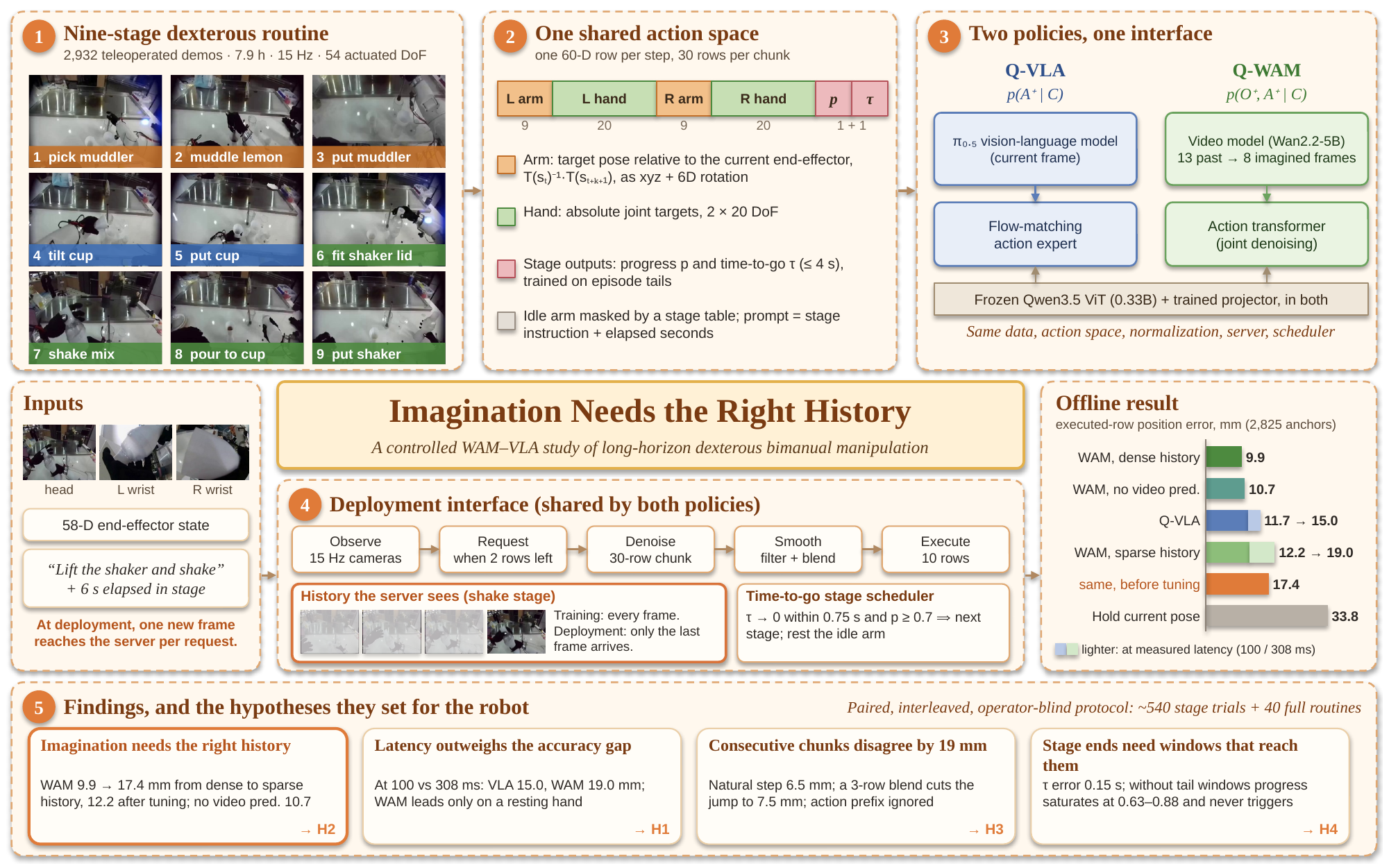}
\caption{\textbf{Overview.} (1)~A nine-stage bimanual routine learned from
2{,}932 glove-teleoperated demonstrations. (2)~One 60-D action space.
(3)~Two policies that differ in whether future video is modeled. (4)~One
deployment interface: at deployment the server receives one new frame per
request. (5)~Findings and the hypotheses they set. Right: offline position
error; lighter bars show the error of the actions executed at each policy's
measured latency.}
\label{fig:overview}
\end{figure*}

\section{Related Work}

\textbf{World-action models.} Video generation has served as a
planner~\cite{du2023unipi}, as pretraining~\cite{wu2024gr1}, as a
representation~\cite{hu2025vpp} and jointly with action
diffusion~\cite{li2025uva,zhu2025uwm}; recent WAMs build on large video
diffusion backbones~\cite{ye2026dreamzero,li2026lingbotva,kim2026cosmospolicy}.
Long-WAM~\cite{huang2026longwam} shows that longer visual context pays off when
the video foundation is pretrained causally and co-designs asynchronous
execution; OpenWAM~\cite{yu2026openwam,wang2026openwam} compares interaction
and pretraining choices under shared protocols;
Fast-WAM~\cite{yuan2026fastwam} asks whether future imagination is needed at
test time. In these studies the VLA baselines are reported benchmark numbers.
Ours is trained on the same dexterous data with the same action space, and the
variable is the history available at deployment, where prior work varies the
context length used in training.

\textbf{Chunked execution and inference delay.} Chunked action
prediction~\cite{zhao2023aloha,chi2023diffusion} underlies current
VLAs~\cite{black2024pi0,pi05,kim2024openvla}. The jump between chunks is
treated by temporal ensembling~\cite{zhao2023aloha}, closed-loop
resampling~\cite{liu2025bid} or real-time chunking~\cite{black2025rtc}, which
also accounts for the actions executed while the next chunk is computed; WAMs
rely on consistent forecasts instead~\cite{huang2026longwam}. We measure what
the delay and the jump cost each policy family on a 54-DoF system.

\textbf{Long-horizon dexterous manipulation.} Hierarchical
VLAs~\cite{shi2025hirobot} and planners~\cite{huang2026longwam} sequence skills
above the policy, and glove and motion-capture systems~\cite{wang2024dexcap}
make dexterous data practical. Our routine is fixed, so we can ask whether a
single policy with a time-to-go output replaces the planner.

\section{A Controlled Testbed}
\label{sec:testbed}

\textbf{Task and data.} The robot has two 7-DoF arms, each with a 20-DoF hand
(54 actuated DoF), a head camera and two wrist cameras ($640{\times}480$,
15\,Hz). The task is a nine-stage
drink-mixing routine (Table~\ref{tab:task}) built from three object chains:
muddler (stages 1--3), cup (4--5) and shaker (6--9). Demonstrations are
recorded one stage at a time by teleoperating the robot itself: instrumented
gloves drive the hands and wrist-worn 6-DoF trackers drive the arms, so every
demonstration is a robot trajectory and needs no retargeting. The data
comprise 2{,}932 episodes and 427{,}792 frames (7.9\,h); the last 5\% of each
stage (147 episodes) is held out.

\begin{table}[t]
\centering
\caption{Stages, active arm and mean length.}
\label{tab:task}
\setlength{\tabcolsep}{3.5pt}
\begin{tabular}{rllrrr}
\toprule
\# & Stage & Arm & Episodes & Frames & Len.\,(s) \\
\midrule
1 & pick muddler   & R    & 272 & 30{,}472 & 7.5 \\
2 & muddle lemon   & R    & 336 & 38{,}660 & 7.7 \\
3 & put muddler    & R    & 259 & 27{,}296 & 7.0 \\
4 & tilt cup       & L    & 366 & 72{,}906 & 13.3 \\
5 & put cup        & L    & 344 & 28{,}147 & 5.5 \\
6 & fit shaker lid & R    & 398 & 65{,}980 & 11.1 \\
7 & shake mix      & both & 285 & 71{,}144 & 16.6 \\
8 & pour to cup    & R    & 395 & 65{,}915 & 11.1 \\
9 & put shaker     & R    & 277 & 27{,}272 & 6.6 \\
\midrule
  & total          &      & 2{,}932 & 427{,}792 & 9.7 \\
\bottomrule
\end{tabular}
\end{table}

\textbf{Shared action space.} The state $s_t\in\mathbb{R}^{58}$ holds, for each
arm, a 9-D end-effector pose $\phi(T)$ (position and the first two rows of the
rotation matrix) and the 20 joint angles of its hand. A policy outputs a
\emph{chunk} of $H{=}30$ \emph{rows}, one per 15\,Hz control step. A row has 60
dimensions. For each arm $i$ it holds the target pose relative to the current
one,
\begin{equation}
a^{\text{arm},i}_{t+k} = \phi\!\left(T^{i}(s_t)^{-1}\,T^{i}(s_{t+k+1})\right),
\end{equation}
for each hand 20 absolute joint targets, and two stage outputs: progress
$p\in[0,1]$ and time-to-go $\tau=\mathrm{clip}(r-k/15,\,0,\,4\,\mathrm{s})$,
where $r$ is the time left in the stage at frame $t$. The two arms and two
hands carry equal loss weights and each stage output a weight of 0.05; the
pose of an arm that a stage leaves idle is masked. The language input is the
stage instruction and the time elapsed in the stage. Training windows are
sampled uniformly over frames and include those that reach into the last 2\,s
of an episode (\emph{tail windows}).

\textbf{Two policies.} Both map the same context $C=(o_{\le t}, s_t, \ell)$ to
a chunk $A^+$ and differ in whether the future observation $O^+$ is modeled,
\begin{equation}
\text{VLA: } p(A^+\mid C), \qquad \text{WAM: } p(O^+, A^+\mid C).
\end{equation}
\vla\ starts from $\pi_{0.5}$~\cite{pi05} and adds a frozen Qwen3.5 vision
tower~\cite{qwen35} (0.33B) with a trained projector; the tower reads the three
camera views tiled into one $384{\times}320$ image. \vla\ sees the current
frame only and is trained for 120k steps at batch 128. \wam\ starts from the
publicly released Long-WAM checkpoint~\cite{huang2026longwam}, a
Wan2.2-TI2V-5B video model coupled with an action transformer, to which we add
the same tower and projector. From 13 past tiled frames spanning 1.6\,s
(\emph{dense history}) it jointly denoises 8 future frames and a chunk. It is
trained for 40k steps on dense history (batch 112) and then fine-tuned for 12k
steps on a mixture of dense and \emph{sparse history}, in which past frames
are 6--12 frames apart as they are at deployment. Unless noted, \wam\ denotes
this fine-tuned model. We call the 40k-step model \emph{dense-trained}, and
the same model trained for 40k steps with no imagined frames and no video loss
the \emph{no-video control}. The deployed setting, used for \wam\ under sparse
history and for all latency results, is 5 action and 1 video denoising steps;
it leaves the accuracy of the dense-trained model unchanged (9.6 vs.\ 9.9\mm\
with the default 10 action steps, which the other models use). Table~\ref{tab:scale} lists what is not matched
between the two policies.

\begin{table}[t]
\centering
\caption{What differs between the two policies (training on H100).}
\label{tab:scale}
\setlength{\tabcolsep}{3pt}
\resizebox{\columnwidth}{!}{%
\begin{tabular}{lll}
\toprule
 & \vla & \wam \\
\midrule
backbone & $\pi_{0.5}$ & Wan2.2-TI2V-5B + action transformer \\
parameters (trained) & 3.5B (3.2B) & 7.1B (2.8B) \\
added vision & \multicolumn{2}{c}{frozen Qwen3.5 ViT 0.33B + trained projector (both)} \\
visual input & current frame & 13 past frames (1.6\,s) \\
training & 120k steps, 315 GPU-h & 40k + 12k steps, 415 GPU-h \\
request latency$^\ast$ & 100\,ms & 308\,ms \\
\bottomrule
\multicolumn{3}{l}{\footnotesize $^\ast$End to end on one RTX\,4090: JPEG encoding, WebSocket round trip, inference.}
\end{tabular}}
\end{table}

\textbf{Shared deployment interface.} Both policies are served through one
WebSocket interface. The robot executes 10 rows of a chunk and requests the
next chunk when 2 rows remain. The server therefore receives one new frame per
request (sparse history), and a reply that arrives $d$ frames after its observation is played
from row $d$. End to end (JPEG encoding of the three views, WebSocket round
trip over loopback, inference) a request takes 100\,ms for \vla\ (p95 103) and
308\,ms for \wam\ (p95 316) on one RTX\,4090, of which encoding and transport
are 2.5\,ms (\vla\ timed on its VR-data variant, which has the same
architecture). Replies thus arrive with a delay of $d{=}2$ and $d{=}5$ frames, with 33 and
25\,ms to spare for the physical network. A rule-based scheduler advances the
stage instruction when two consecutive replies agree that $\tau$ reaches zero
within 0.75\,s and $p\ge0.7$, and returns the arm that becomes idle to its
rest pose. Optionally the server smooths each chunk with a Savitzky--Golay
filter (order 2, window 9) and blends its first 3 rows into the previous
chunk.

\section{Preliminary Findings}
\label{sec:findings}

\textbf{Protocol.} On the held-out episodes we query each policy every 5
frames (2{,}825 \emph{anchors}) and compare its rows with the recorded future
as absolute end-effector targets; stages are weighted equally. Three
quantities follow the deployment interface. The \emph{executed-row error}
$e(d)$ is the mean error of rows $d,\dots,d{+}9$, the rows played when the
reply has delay $d$; $e(0)$, zero delay, is what offline evaluation
conventionally reports and no deployed policy attains. The \emph{seam} is the
distance between the last row executed from one chunk and the first row
executed from the next. The \emph{history} is dense or sparse as defined
above. We report position and rotation error of the active arm and hand-joint
error. Intervals are 95\% bootstrap intervals over episodes, resampled within
each stage (2{,}000 replicates); differences are paired on the same episodes.

\begin{table}[t]
\centering
\caption{Executed-row error (stage-balanced; 95\% CI for position). Bold marks
the better policy of a pair. Seam and time-to-go error are at zero
delay.}
\label{tab:main}
\setlength{\tabcolsep}{3pt}
\resizebox{\columnwidth}{!}{%
\begin{tabular}{lccccc}
\toprule
Policy & pos (mm) & rot ($^\circ$) & hand ($^\circ$) & seam (mm) & $\tau$ err (s) \\
\midrule
\multicolumn{6}{l}{\emph{sparse history, zero delay}} \\
\vla & \textbf{11.7} {\scriptsize[11.3, 12.2]} & 3.37 & 1.15 & 19.6 & 0.16 \\
\wam & 12.2 {\scriptsize[11.9, 12.6]} & \textbf{3.24} & \textbf{0.82} & 19.5 & 0.18 \\
\midrule
\multicolumn{6}{l}{\emph{sparse history, measured delay}} \\
\vla\ ($d{=}2$) & \textbf{15.0} {\scriptsize[14.4, 15.6]} & \textbf{4.29} & 1.25 & -- & -- \\
\wam\ ($d{=}5$) & 19.0 {\scriptsize[18.4, 19.7]} & 5.21 & \textbf{1.13} & -- & -- \\
\midrule
\multicolumn{6}{l}{\emph{\vla\ variants, zero delay}} \\
+ 693 VR-teleoperated routines$^\dagger$ & 11.6 {\scriptsize[11.2, 12.0]} & 3.35 & 1.14 & 19.3 & 0.15 \\
\quad + 2 past frames & 11.7 {\scriptsize[11.3, 12.2]} & 3.38 & 1.15 & 19.5 & 0.15 \\
\midrule
\multicolumn{6}{l}{\emph{references, zero delay}} \\
dense-trained \wam, dense history & 9.9 {\scriptsize[9.6, 10.2]} & 2.79 & 0.80 & 17.6 & 0.18 \\
hold current pose & 33.8 {\scriptsize[33.3, 34.5]} & 7.80 & 0.97 & -- & -- \\
\bottomrule
\multicolumn{6}{l}{\footnotesize $^\dagger$\vla\ fine-tuned on the glove data plus 693 full routines teleoperated in VR.}
\end{tabular}}
\end{table}

\subsection{What the WAM sees}

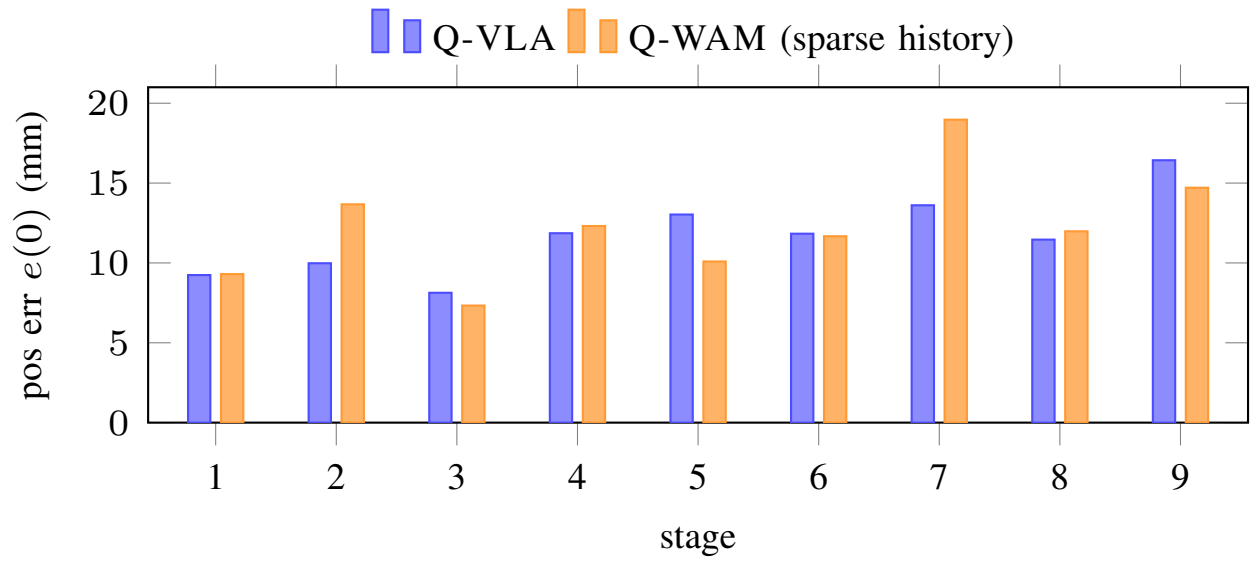
\begin{figure}[t]
\centering
\begin{tikzpicture}
\begin{axis}[
  ybar, bar width=4.2pt, width=\columnwidth, height=38mm,
  ylabel={pos err $e(0)$ (mm)}, ylabel style={font=\scriptsize},
  symbolic x coords={1,2,3,4,5,6,7,8,9},
  xtick=data, xlabel={stage}, xlabel style={font=\scriptsize},
  ticklabel style={font=\scriptsize}, ymin=0, ymax=21,
  legend style={font=\scriptsize, at={(0.5,1.02)}, anchor=south, draw=none, fill=none},
  legend columns=2, enlarge x limits=0.07]
\addplot+[fill=blue!45, draw=blue!70] coordinates {(1,9.24) (2,9.98) (3,8.13) (4,11.86) (5,13.03) (6,11.83) (7,13.61) (8,11.46) (9,16.43)};
\addplot+[fill=orange!60, draw=orange!80] coordinates {(1,9.30) (2,13.67) (3,7.33) (4,12.31) (5,10.09) (6,11.67) (7,18.97) (8,11.98) (9,14.71)};
\legend{\vla, \wam\ (sparse history)}
\end{axis}
\end{tikzpicture}
\caption{Per-stage position error with sparse history at zero delay.
\wam\ loses most on the periodic stages 2 (muddle) and 7 (shake).}
\label{fig:stage}
\end{figure}

\finding{At zero delay and with sparse history, neither policy dominates.} \vla\ is
more accurate in position by 0.49\mm\ (CI 0.13--0.85), \wam\ on hand joints by
0.33$^\circ$ (CI 0.30--0.36, 29\%) and marginally in rotation (0.13$^\circ$, CI
0.00--0.25) (Table~\ref{tab:main}). \wam\ leads on the placing stages 5 and 9
and trails on muddling and shaking (Fig.~\ref{fig:stage}).

\finding{Dense history is worth 2.3\mm\ to the WAM.} With dense history the
dense-trained model reaches 9.9\mm, 1.9\mm\ ahead of \vla\ (CI 1.5--2.2), and
\wam\ reaches the same 9.9\mm. Given sparse history, the dense-trained model
degrades to 17.4\mm, and to 31.6\mm\ when it receives the current frame only,
as on the first request of a stage (Table~\ref{tab:abl}). Fine-tuning on
sparse history reduces these errors to 12.2 and 19.0\mm\ at no cost under
dense history. The periodic stages stay behind (muddling 13.7\mm, shaking
19.0\mm), because the phase of a fast periodic motion cannot be read from
frames 0.67\,s apart.

\begin{table}[t]
\centering
\caption{Position error $e(0)$ in mm of three WAM models under three
histories at inference.}
\label{tab:abl}
\setlength{\tabcolsep}{4pt}
\resizebox{\columnwidth}{!}{%
\begin{tabular}{lccc}
\toprule
History at inference & dense-trained & \wam\ (fine-tuned) & no-video control \\
\midrule
dense (13 frames, 1.6\,s) & 9.9 & 9.9 & 10.7 \\
sparse (one frame per request) & 17.4 & 12.2 & 20.4 \\
current frame only & 31.6 & 19.0 & 30.9 \\
\bottomrule
\end{tabular}}
\end{table}

\finding{Video prediction helps, most when history is sparse.} With dense
history the no-video control reaches 10.7\mm: 0.81\mm\ behind the dense-trained
model (CI 0.62--1.01) and 1.04\mm\ ahead of \vla\ (CI 0.71--1.36). Hand-joint
error falls in the same order, 1.15, 0.93 and 0.80$^\circ$. Under sparse
history the gap between the two dense-trained models grows to 2.97\mm\ (CI
2.66--3.30), so imagined frames partly stand in for missing history. The
control keeps the video-pretrained backbone; the comparison therefore isolates
the video objective and the imagined frames and leaves the value of video
pretraining open. One video denoising step suffices
(Sec.~\ref{sec:testbed}), so the imagined future can be coarse.

\finding{More data helps the VLA little, two past frames not at all.} Adding
693 full routines teleoperated in VR improves \vla\ by 0.14\mm\ (CI
0.05--0.25). Adding two past
frames at request spacing, encoded by the frozen tower with a learned
per-frame embedding and trained for 30k steps, makes that variant worse by
0.16\mm\ (CI 0.11--0.22). A \vla\ with the dense history that \wam\ sees is
untested and planned for the full paper.

\subsection{When the reply arrives}

\begin{figure}[t]
\centering
\begin{tikzpicture}
\begin{axis}[
  width=\columnwidth, height=40mm,
  xlabel={reply delay $d$ (frames at 15\,Hz; one frame is 67\,ms)},
  ylabel={pos err $e(d)$ (mm)},
  xlabel style={font=\scriptsize}, ylabel style={font=\scriptsize},
  ticklabel style={font=\scriptsize}, xmin=-0.3, xmax=8.3, ymin=8, ymax=25.5,
  xtick={0,...,8}, ymajorgrids, grid style={gray!25},
  legend style={font=\scriptsize, at={(0.5,1.03)}, anchor=south, draw=none, fill=none, /tikz/every even column/.append style={column sep=5pt}},
  legend columns=3]
\addplot[blue!70, thick, mark=*, mark size=1.2pt] coordinates {(0,11.73) (1,13.40) (2,14.98) (3,16.49) (4,17.91) (5,19.23) (6,20.46) (7,21.61) (8,22.69)};
\addplot[orange!85!black, thick, dashed, mark=square*, mark size=1.2pt, mark options={solid}] coordinates {(0,12.22) (1,13.75) (2,15.21) (3,16.58) (4,17.85) (5,19.03) (6,20.13) (7,21.19) (8,22.22)};
\addplot[green!45!black, thick, densely dotted, mark=triangle*, mark size=1.5pt, mark options={solid}] coordinates {(0,9.88) (1,11.30) (2,12.72) (3,14.16) (4,15.56) (5,16.88) (6,18.14) (7,19.35) (8,20.51)};
\legend{\vla, \wam\ (sparse), dense-trained (dense)}
\draw[blue!70, thick] (axis cs:2,14.98) circle[radius=3.2pt];
\draw[orange!85!black, thick] (axis cs:5,19.03) circle[radius=3.2pt];
\draw[blue!70, thin] (axis cs:2,15.7) -- (axis cs:2,18.6);
\node[font=\scriptsize, anchor=south, inner sep=1pt, text=blue!70!black] at (axis cs:2,18.6) {\vla\ at 100\,ms};
\draw[orange!85!black, thin] (axis cs:5,19.75) -- (axis cs:5,22.4);
\node[font=\scriptsize, anchor=south, inner sep=1pt, text=orange!60!black] at (axis cs:5,22.4) {\wam\ at 308\,ms};
\end{axis}
\end{tikzpicture}
\caption{Executed-row position error against reply delay, for \wam\ with
sparse history and the dense-trained model with dense history. The \vla\ and
sparse-history curves coincide;
circles mark each policy at its measured end-to-end latency, 4.0\mm\ apart.}
\label{fig:lat}
\end{figure}
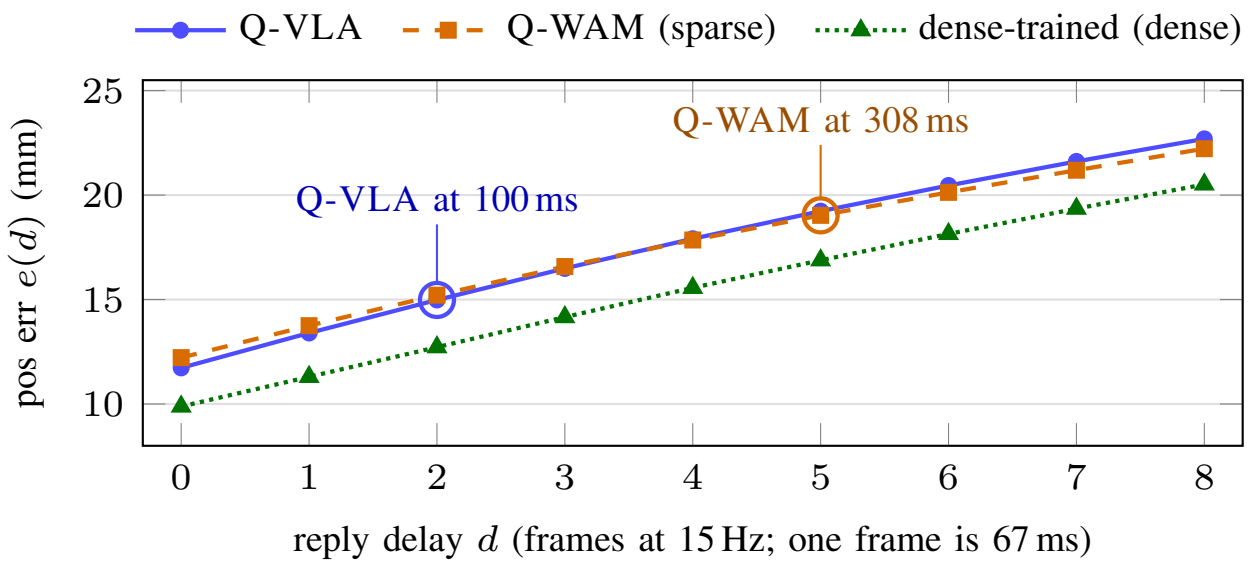

\finding{At the measured latency the VLA leads by 4.0\mm.} Each frame of delay
adds 1.4--1.5\mm\ to the executed-row error of either policy
(Fig.~\ref{fig:lat}). At their measured delays \vla\ is at 15.0\mm\ and \wam\
at 19.0\mm, a difference of 4.0\mm\ (CI 3.6--4.5) and eight times the zero-delay
gap. Even the dense-trained model with dense history stays behind (16.9\mm,
1.9\mm, CI 1.4--2.4). At equal delay the two policies stay within 0.5\mm\ of each other
for every $d\le8$ (0.24\mm\ at $d{=}2$, CI $-$0.23--0.69). The deficit
therefore comes from the extra 200\,ms, and \wam\ would tie only if it replied
within 133\,ms. On hand joints \wam\ keeps a lead of 0.11$^\circ$ (CI
0.08--0.15).

\finding{The WAM's hand lead comes from a steadier resting hand.} We split the
anchors by the recorded hand speed over rows 0--9. In the slower half (below
1.9$^\circ$/s; 1{,}412 anchors) holding the current pose has an error of
0.11$^\circ$, \vla\ 1.03$^\circ$ and \wam\ 0.52$^\circ$. Both policies move a
hand that should rest, \vla\ twice as much, and \wam's lead persists at the
measured latency (0.38$^\circ$, CI 0.31--0.44). In the fastest quartile (above
7.1$^\circ$/s; grasps and releases) the zero-delay lead is 0.15$^\circ$ (CI
0.07--0.24, 9\%), and at the measured latency \vla\ is ahead by 0.19$^\circ$
(CI 0.08--0.30) and by 4.8\mm. The position lead of the dense-trained model
under dense history also disappears in this quartile (0.3\mm, CI
$-$0.1--0.8). Anchor-weighted averages give the same signs.

\subsection{How chunks and stages join}

\finding{Consecutive chunks disagree by 19\mm.} At zero delay the seam
is 19.3--19.6\mm\ for every policy, and 20.5--22.1\mm\ at $d{=}5$, against a
frame-to-frame step of 6.5\mm\ in the demonstrations. \wam, whose chunks come
from a continuous video rollout, is no smoother than \vla\ (difference
$-$0.1\mm, CI $-$0.8--0.5); only dense history lowers the seam (17.6\mm\ for
the dense-trained model). The
seam contains the errors of both chunks and is measured on commands, so the
robot's controller may filter part of it. Replaying logged chunks, a 3-row
blend reduces the seam to 7.3--7.9\mm\ at a cost of 1.0--1.6\mm\ in
executed-row error, and Savitzky--Golay smoothing halves \wam's command jerk
at unchanged accuracy. Training \wam\ with the rows already committed as an
input does not help: executed-row error (19.1 vs.\ 19.0\mm) and seam (21.5
vs.\ 23.2\mm) barely change.

\finding{Stage ends are learned only from windows that reach them.} Without
tail windows the predicted progress saturates at 0.63--0.88 and a progress
threshold never fires. With tail windows, time-to-go is predicted to within
0.15\,s (\vla) and 0.18\,s (\wam; 0.27--0.29\,s on muddling). Whether
time-to-go switches stages better than progress trained on the same windows
has not been tested yet (H4).

\section{Hypotheses and Real-Robot Protocol}
\label{sec:hyp}

The findings set four hypotheses (Table~\ref{tab:hyp}). Each policy runs at its
own latency. Tests are one-sided at $\alpha{=}0.05$ with Holm adjustment over
the five primary endpoints; the power figures below are unadjusted.

\begin{table}[t]
\centering
\caption{Hypotheses, stages, paired trials and primary endpoints.}
\label{tab:hyp}
\setlength{\tabcolsep}{3pt}
\resizebox{\columnwidth}{!}{%
\begin{tabular}{llcl}
\toprule
 & Stages & Trials & Primary endpoint (test) \\
\midrule
H1a no hand lead: \vla\ succeeds & 1, 3, 6 & 60 & stage success (exact McNemar) \\
H1b hand lead: \wam\ keeps the object & 8, 9 & 60 & object lost (exact McNemar) \\
H2 dense history restores phase & 2, 7 & 30 & stroke-period deviation (Wilcoxon) \\
H3 blending helps & 7--9 & 60 & stage success, measured jerk \\
H4 time-to-go chains stages & 1--9 & 40$^\ddagger$ & stages completed per routine \\
\bottomrule
\multicolumn{4}{l}{\footnotesize $^\ddagger$Full routines: 20 time-to-go, 10 progress threshold, 10 oracle switching.}
\end{tabular}}
\end{table}

\textbf{H1: the hand lead matters where it exists.} At the measured latency
\wam\ trails in position on every stage but one. (a)~On the three stages where
it also has no hand lead (1, 3, 6), \vla\ succeeds more often. (b)~On the two
stages where its hand lead is largest (8, 9; 0.3--0.6$^\circ$), \wam\ loses the
object less often (drop, slip or spill). Sixty pooled paired trials per part
give 83--90\% power for a 25-point difference and 65--72\% for 20 points
(simulated for base rates 0.4--0.6). Two controls of 60 trials each separate
the causes: the no-video control (imagination) and \vla\ with its replies
delayed to $d{=}5$ (latency).

\textbf{H2: dense history restores phase.} Streaming dense history from the
robot at 15\,Hz brings \wam's stroke period on stages 2 and 7 closer to the
demonstrations'. The endpoint is the deviation of the executed stroke period,
computed from logged end-effector poses; 30 pooled paired trials give 80\%
power for an effect of 0.5 standard deviations. Success rate is secondary,
because 30 trials reach 80\% power only above about 35 points. Streaming
requires encoding frames while the robot executes, as
in~\cite{huang2026longwam}; we report the added latency.

\textbf{H3: blending helps at the seam.} A 3-row blend raises success and
lowers measured jerk on stages 7--9 for both policies, despite its
1.0--1.6\mm\ offline cost.

\textbf{H4: time-to-go chains stages.} With the time-to-go scheduler, the
policy with the higher pooled success in H1 completes more stages per routine
than with a progress threshold trained on the same tail windows, and at most
one stage fewer than with oracle switching. Both rules are first compared by
replaying logged predictions.

\textbf{Further baselines.} $\pi_{0.5}$ on the same nine stages without the
Qwen tower; \vla\ with the dense history that \wam\ sees; nine single-stage
$\pi_{0.5}$ experts with oracle switching as a chaining reference.

\textbf{Protocol.} Object poses come from a fixed set shared by all policies,
policies are interleaved in randomized blocks, the operator does not know
which server runs, and a second person judges from video. A stage succeeds if
its goal state (for example muddler back in its holder, lid seated, tea in the
cup) holds within twice the mean demonstration length, without intervention
and without a dropped or toppled object; goal states are fixed before the
trials. We report Wilson 95\% intervals and a stage$\times$policy logistic
regression. H1--H4 need about 540 stage trials and 40 full routines.

\section{Discussion and Limitations}

\textbf{What the offline evidence shows.} With identical data and action
space, the WAM's lead over the VLA is real when history is dense and delay is
zero, and part of it is due to video prediction. Once the history is
what a robot sends and the reply arrives when it actually does, the lead is
confined to a hand at rest. On this task and hardware the extra 200\,ms of
inference costs more (4.0\mm) than dense history gains (2.3\mm) or video
prediction gains (0.8\mm). We therefore suggest
that WAM--VLA comparisons report the error of the actions executed at the
measured latency, under deployable history, next to zero-delay error.

\textbf{What it cannot show.} Open-loop action error is an imperfect proxy for
success: a closed loop may absorb a 4\mm\ arm error or amplify a 0.2$^\circ$
hand error, and H1 is built to tell which. If H1b and H2 hold, the case for
WAMs on dexterous hands is real but conditional on streamed history and on
replying about as fast as the VLA. If the robot ranks the policies differently
from our offline protocol, that is a result about offline evaluation in
itself.

\textbf{Limitations.} The study covers one platform and one task family. The
two policies differ in size, training budget and latency, which we report but
do not equalize, and \vla\ was not given dense history. Latency was measured on
one GPU without the physical network. The demonstrations are teleoperated
robot trajectories, so offline error omits the contact and tracking errors of
closed-loop execution.

\end{document}